\newcommand{\D}{\mathcal{D}}

\newcommand{\bw}{\mathbf{\Theta}}
\newcommand{\bwj}{\mathbf{\Theta}_{j}}
\newcommand{\bwo}{\mathbf{\Theta}_{\textrm{o}}}
\newcommand{\bwoj}{\mathbf{\Theta}_{\textrm{o}, j}}
\newcommand{\w}{\theta}

\newcommand{\nw}{D}
\newcommand{\ny}{M}
\newcommand{\na}{n_a}
\newcommand{\nb}{n_b}

\newcommand{\Q}{Q}

\documentclass[letterpaper,11pt]{article}

\usepackage{graphicx} 
\usepackage{amsmath}
\usepackage{amsfonts}
\usepackage{amssymb}
\usepackage{xcolor}
\usepackage{subcaption}
\usepackage{hyperref} 

 \title{Gradient-based bilevel optimization for multi-penalty Ridge regression through matrix differential calculus} %
\author{Gabriele Maroni, Loris Cannelli, Dario Piga}

\date{
IDSIA Dalle Molle Institute for Artificial Intelligence USI-SUPSI, Via la Santa 1, CH-6962 Lugano-Viganello, Switzerland.\\
\ \\
November 23 2023}

\begin{document}

\maketitle

%

%




\begin{abstract}                          
Common regularization algorithms for linear regression, such as LASSO and Ridge regression, rely on a regularization hyperparameter that balances the tradeoff between minimizing the fitting error and the norm of the learned model coefficients. As this hyperparameter is scalar, it can be easily  selected via random or grid search optimizing a cross-validation criterion. However, using a scalar hyperparameter limits the algorithm's flexibility and potential for better generalization. In this paper, we address the problem of linear regression with  $\ell_2$-regularization, where a different regularization hyperparameter is associated with each input variable. We optimize these hyperparameters using a gradient-based approach, wherein the gradient of a cross-validation criterion with respect to the regularization hyperparameters is computed analytically through matrix differential calculus. Additionally, we introduce two strategies tailored for sparse model learning problems aiming at reducing the risk of overfitting to the validation data. Numerical examples demonstrate that our multi-hyperparameter regularization approach outperforms LASSO, Ridge, and Elastic Net regression. Moreover, the analytical computation of the gradient proves to be more efficient in terms of computational time compared to automatic differentiation, especially when handling a large number of input variables. Application to the identification of over-parameterized Linear Parameter-Varying models is also presented.
 \end{abstract}

 \section{Introduction}

\subsection{Research problem}

Linear regression problems often require regularization of the model parameters for two main reasons:
($i$) to ensure better numerical conditioning in cases of multicollinearity (i.e., when input variables are highly correlated);
($ii$) to enhance the generalization capabilities of the estimated model, as regularization  drives the model coefficients towards zero, thereby reducing model complexity and preventing overfitting.

The most widely used regularization techniques for linear regression are the LASSO \cite{tibshirani1996regression} and the Ridge algorithm \cite{hoerl1970Ridge}. These techniques penalize the 
$\ell_1$ and the $\ell_2$  norm of the model coefficients, respectively. In the simplest case, the importance of the penalty term is governed by a scalar hyperparameter, which controls the balance between fitting the training data and enforcing a penalty on the norm of the model coefficients. The use of a singular scalar hyperparameter simplifies its selection, usually accomplished through random or grid search within a cross-validation framework.  
However, relying solely on a single hyperparameter presents certain limitations. Specifically, it applies the same regularization strength to all input variables without considering the varied importance or relevance of different features. Furthermore, regularization inherently introduces bias by shrinking model coefficients towards zero, and relying on just one hyperparameter limits the flexibility in managing the bias-variance tradeoff. In cases involving inherently sparse models, where only a few input variables are relevant, it may be more suitable to encourage only the model parameters associated with non-relevant variables to approach zero, rather than  penalizing the other model coefficients as much or at all.

While addressing these limitations by introducing multiple hyperparameters, potentially assigning one to each input feature, seems theoretically appealing, it comes with challenges. The curse of dimensionality becomes apparent, and the risk of overfitting increases. The search space for hyperparameters grows significantly, making grid search or random search strategies practically infeasible. Even more scalable gradient-free algorithms, such as Bayesian optimization \cite{snoek2012practical}, are generally limited to handling a maximum of roughly  50 hyperparameters. Paradoxically, while regularization aims to prevent overfitting, introducing numerous hyperparameters, such as a regularization coefficient for each feature, can inadvertently lead to overfitting the validation set during hyperparameter tuning.

\subsection{Contribution}

This paper focuses on linear regression with an $\ell_2$ penalty on the norm of the model coefficients, employing multiple hyperparameters. We consider the general scenario where each input variable is associated with its unique hyperparameter. Essentially, our problem can be seen as a generalization of Ridge regression with multiple hyperparameters. Throughout this  paper, we will refer to this problem  as \emph{multi-ridge regression}, adopting the terminology in~\cite{van2021fast}.   
To select these hyperparameters, we employ a gradient-based approach that optimizes a cross-validation criterion. The gradient  with respect to the regularization hyperparameters is analytically computed using  \emph{matrix differential calculus}.  This computation leverages efficient matrix-vector multiplication available in modern scientific computing libraries such as NumPy, PyTorch, and JAX.

As outlined in \cite{bengio2000gradient}, optimizing a large number of hyperparameters through cross-validation can lead to overfitting to the validation set. To mitigate this risk, we introduce two strategies that are particularly suitable for solving sparse model learning problems.

In the numerical examples, we will demonstrate that using multiple regularization hyperparameters enhances the performance compared to single-hyperparameter Ridge regression. Furthermore, it outperforms both LASSO and Elastic Net, even in the context of sparse model learning problems.

To ensure the reproducibility of our results and to encourage further contributions to the field, we have made a PyTorch implementation of our proposed approach.  This implementation is tailored with a scikit-learn compatible interface, ensuring seamless integration into existing workflows. Both the implementation and the results presented in this paper are freely available on the GitHub repository at the following link: \href{https://github.com/gabribg88/Multiridge}{https://github.com/gabribg88/Multiridge}. 

\subsection{Related works}
The problem of Ridge regression with multiple penalties was early proposed in~\cite{hoerl1970Ridge}. 
A gradient-based approach to optimize the algorithm's hyperparameter was later addressed  in the seminal work~\cite{bengio2000gradient}.  The author here proposes to solve the inefficient step of computing the gradient of the problem variables with respect to the hyperparameters through a combination of Cholesky decomposition and backpropagation. The approach introduced in \cite{bengio2000gradient}, later implemented in \cite{latendresselinear},  has been then extensively explored in subsequent studies. For instance,  the works \cite{franceschi2017forward,hataya2022meta,lorraine2020optimizing,pedregosa2016hyperparameter,bertrand2022implicit} propose efficient solutions to compute the computationally expensive gradient step mentioned in \cite{bengio2000gradient}. These solutions employ approximators (e.g., Neumann series as demonstrated in \cite{lorraine2020optimizing,hataya2022meta}) or iterative procedures (e.g., conjugate gradient in \cite{pedregosa2016hyperparameter}, reverse-hyper gradient in \cite{franceschi2017forward}, or proximal non-smooth techniques in \cite{bertrand2022implicit}). Although these approaches provide approximate solutions, they have proven to be satisfactory, particularly for machine learning tasks, as discussed \cite{pedregosa2016hyperparameter,blondel2022efficient}.

Another research avenue that has gained momentum from \cite{bengio2000gradient} is the exploration of optimizing regularization hyperparameters by targeting specific machine-learning problems. By selecting a particular machine learning application, researchers can frame the task of finding the optimal variable fitting, while simultaneously determining the best configuration for the hyperparameters, as a bilevel optimization problem. This allows for tailored approaches to effectively address the challenge. Interesting tools have been proposed, among others, in \cite{bennett2006model} for regularized Support Vector Machines, in \cite{foo2007efficient} for log-linear models, in \cite{frecon2018bilevel} for the group-LASSO setting, and in \cite{kunisch2013bilevel} for Ridge and LASSO regularizations. An alternative approach to  nested optimization for selecting the multi-ridge hyperparameters is based on maximization of the marginal likelihood and discussed  in \cite{van2021fast}. 

In relation to existing literature, we present a custom approach to linear regression with $\ell_2$ regularization. As previously highlighted, this approach enables us to  compute efficiently an analytical expression for the gradient of a cross-validation-based criterion with respect to the hyperparameters, using matrix differential calculus.

\subsection{Paper organization}

The paper is organized as follows. Section~\ref{Sec:notation} introduces the mathematical notation used throughout the paper. Section~\ref{Sec:probform} presents the multi-hyperparameter regression problem addressed in our contribution, along with an intuitive motivation of the advantages of using multiple regularization hyperparameters over a single one, especially in sparse model learning problems. In Section~\ref{Sec:approach}, we present the analytical computation of the gradient of the cross-validation criterion with respect to the regularization hyperparameters. Detailed derivations can be found in the Appendix to facilitate a smooth reading flow. Section~\ref{sec:data_aug} discusses the two strategies we propose to reduce the risk of overfitting to the validation set. Section \ref{Sec: example} provides numerical examples showcasing the effectiveness of our approach. One of these examples is focused on the identification of over-parameterized Linear Parameter-Varying models, a topic previously tackled in the literature using LASSO-like and kernel-based algorithms (see, e.g., \cite{toth2012order,piga2013lpv,laurain2020sparse}). Finally, Section~\ref{Sec:conclusion} offers conclusions and directions for future research.

\section{Notation} \label{Sec:notation}
We denote with  $\otimes$ the Kronecker product and with  $\odot$ the elementwise matrix multiplication.  
Given a column vector $v \in \mathbb{R}^n$, 
$\mathrm{Diag}(v)$ is an  $n \times n$ diagonal matrix having the elements of $v$ itself on the main diagonal and 0s otherwise; conversely, given a matrix $V \in \mathbb{R}^{n \times n}$,  $\mathrm{diag}(V)$  is the column vector $v\in\mathbb{R}^n$ collecting the 
entries of the main diagonal of $V$. $\mathrm{vec}(\cdot)$ denotes  the vectorization operator that stacks the columns of a matrix   vertically to form a column vector. Given any matrix $A\in\mathbb{R}^{m\times n}$, $\left\| A \right\|_{F}$ denotes its Frobenius norm, and $A'$ the transpose of $A$. Finally, we indicate with   $I_N$ the identity matrix of size $N$.

Given a matrix  $U \in \mathbb{R}^{m \times n}$ and a real-valued function $\phi: \mathbb{R}^{m \times n} \rightarrow \mathbb{R}$, $U \mapsto \phi(U)$, the gradient of $\phi$ w.r.t. the vectorized version  of $U$ ($\mathrm{vec}(U)$) is denoted as $\nabla_{U}\phi$, with $\nabla_{U}\phi \in \mathbb{R}^{mn}$. If  $U \in \mathbb{R}^{m \times m}$ is a diagonal matrix, with diagonal elements $u = \mathrm{diag}(U) \in \mathbb{R}^m$,  $\nabla_{u}\phi \in \mathbb{R}^m$ is the gradient of   the function $\phi$ w.r.t. the diagonal elements of the matrix $U$.

Given a finite-dimensionals set $\mathcal{S}$, $|\mathcal{S}|$ is its cardinality.

\section{Problem Formulation} \label{Sec:probform}

\subsection{Parametric estimate: setting}
In a parametric supervised learning setting, we aim to select a function $f$ that maps input features $x \in \mathcal{X}$ to output targets $y \in \mathcal{Y}$ from a set $\mathcal{F}$ of candidate functions, parameterized by a vector of parameters $\w$, such that each value assumed by the vector of parameters $\w$ corresponds to a different function in the set $\mathcal{F}$. In this work we focus on regression problems, and we assume $\mathcal{X} = \mathbb{R}^D$, with $D$ being the number of input features; and $\mathcal{Y} = \mathbb{R}$ in the case of simple regression with a real-valued target, or $\mathcal{Y} = \mathbb{R}^M$ in the case of multi-task/target regression, with $M$ denoting the number of real-valued target variables. 

In a real-world setting, we are given a dataset of $N$ input-output pairs $\mathcal{D}=\left\{\left(x^{(i)}, y^{(i)}\right)\right\}_{i=1}^N$ which are assumed to be independently drawn from the same unknown probability distribution $\mathbb{P}_{x,y}$. In principle, we would like to select the values of $\w$ (and hence the function $f_{\w} \in \mathcal{F}$) which minimize the expected risk,  defined as the expected value $\mathbb{E}_{x,y}[\ell(y, f_{\w})]$ of some scalar loss function $\ell(y, f_{\w})$, which is a measure of the discrepancy between the target and the model's prediction. 
In general we do not have access to the data distribution $\mathbb{P}_{x,y}$, so we approximate the expected risk with an empirical risk defined as the average loss over the dataset $\D$:

\begin{equation}
    L(\w) =  \frac{1}{N} \sum_{i=1}^N \ell\left(y^{(i)}, f_{\w}\left(x^{(i)}\right)\right).
\end{equation}

In solving regression problems, a regularization term $R(\w, \lambda)$ is typically added to control the complexity of the estimated function $f_{\w}$. Such a regularization term is parameterized by a vector of hyperparameters $\lambda$  that regulate the trade-off between the fit  and the regularization term.

The overall training criterion $C(\w, \lambda)$ is then defined as:

\begin{equation} \label{eqn:train_criterion}
    C(\w, \lambda) =  L(\w) + R(\w, \lambda),
\end{equation}
    
and, for a given value of $\lambda$, the problem   of finding the optimal parameterized function is typically formulated as the following optimization problem:

\begin{equation} \label{eqn:inner_optimization}
    \hat{\w}(\lambda) = \arg\min_{\w} C(\w, \lambda).   
\end{equation}

Among different possibilities, the hyperparameters $\lambda$ can be chosen by optimizing an evaluation criterion $E(\hat{\w}(\lambda))$, such as minimizing the average loss over a new dataset that was not used in evaluating $L(\w)$. For example, in the celebrated $K$-fold cross-validation strategy, the original dataset $\D$ is divided into $K$ partitions $\D_1, \ldots,\D_K$, and, consequently, the evaluation criterion takes the form:

\begin{align} \label{eqn:eval_criterion}
E \left(\hat \w(\lambda)\right) = \frac{1}{K}\sum_{k=1}^K L_k \left(\hat \w^{(\backslash  k)}(\lambda)\right),
\end{align}

where $L_k$ is the average loss over the subset $\D_k$ (namely, \textit{validation fold}). Specifically: 
 
\begin{align} \label{eqn:aver_loss}
L_k\left(\hat \w^{(\backslash  k)}(\lambda)\right) = \frac{1}{|\D_k|} \sum_{(x^{(i)},y^{(i)})  \in \D_k} \ell\left(y^{(i)}, f_{\w^{(\backslash  k)}(\lambda)}\left(x^{(i)}\right)\right),
\end{align}
and 
$\hat \w^{(\backslash k)}(\lambda)$ represents the estimate of the model parameters $\w$ obtained from  constructing $C(\w, \lambda) $ using all remaining folds  $\D_j$ (with $j=1,\ldots,K$ and  $j \neq k$), i.e.,

\begin{align} \label{eqn:Kfold_general}
    &\hat \w^{(\backslash k)}(\lambda) =\\&\arg \min_{\w} \frac{1}{|\bigcup_{j \neq k} \D_{j}|}\sum\limits_{\substack{\left(x^{(i)},y^{(i)}\right) \\ \in \bigcup_{j \neq k} \D_{j} }}  \ell\left(y^{(i)}, f_{\w}\left(x^{(i)}\right)\right) +    R(\w, \lambda).\nonumber
\end{align}

Thus, the hyperparameters  $\lambda$ are selected by solving the minimization problem:

\begin{align} \label{eqn:outer_optimization}
\lambda^{\star} = \arg\min_{\lambda}E \left(\hat \w (\lambda)\right).
\end{align}

The hyperparameter optimization problem is thus a bilevel optimization problem, wherein the objective function of the outer optimization problem \eqref{eqn:outer_optimization} depends on  the solution of the inner optimization task \eqref{eqn:inner_optimization}.

\subsection{Linear-in-the parameter models and quadratic loss}

In this work, we consider:
\begin{itemize}
    \item as the set of candidate functions, linear-in-the-parameter functions of the form
\begin{align}
f_{\bw}(x) =   x' \bw,\label{eqn:choice1}
\end{align}
where $\bw \in \mathbb{R}^{D\times M}$ is the coefficient matrix and $f_{\bw}(x)$ is a row vector of size $\ny$  whose its $i$-th component represents the model prediction for the $i$-th output; 
\item a quadratic fitting loss:
 \begin{align}
\ell\left(y^{(i)}, f_{\bw}\left(x^{(i)}\right)\right) =  \frac{1}{2}\left\| y^{(i)} - \left(x^{(i)}\right)^{'} \bw \right\|^2;\label{eqn:choice2}
 \end{align}
 \item a quadratic regularization term:
 \begin{align}
R(\bw, \lambda) = \frac{1}{2}\left\|  \Lambda \bw \right\|_F^2, \label{eqn:choice3}
 \end{align}
 where $\Lambda = Diag(\lambda)$, and $\lambda \in \mathbb{R}^D$ is the vector of hyperparameters. 
\end{itemize}

Thus,    the general  expressions  \eqref{eqn:Kfold_general} and \eqref{eqn:outer_optimization}  of the $K$-fold cross-validation specialized, respectively, for the above choices \eqref{eqn:choice1}-\eqref{eqn:choice2}-\eqref{eqn:choice3}, become:
\begin{align} \label{eqn:thetaLS1}
    &\hat \bw^{(\backslash k)}(\Lambda) =\\&  \arg \min_{\bw} \frac{1}{2|\bigcup_{j \neq k} \D_{j}|}   \sum\limits_{\substack{\left(x^{(i)},y^{(i)}\right) \\ \in \bigcup_{j \neq k} \D_{j} }} \hspace{-.1cm}  \left\| y^{(i)} - \left(x^{(i)}\right)^{'}\hspace{-.2cm} \bw \right\|^2\hspace{-.2cm}  + \hspace{-.05cm}\frac{1}{2}\hspace{-.1cm} \left\|  \Lambda \bw \right\|_F^2\hspace{-.06cm},\nonumber
\end{align}
and 
\begin{align} \label{eqn:lambdastar1}
&\Lambda^{\star} =\\& \arg\min_{\Lambda} \frac{1}{2K}\sum_{k=1}^K \frac{1}{|\D_{k}|} \sum\limits_{(x^{(i)},y^{(i)}) \in \D_k} \hspace{-.2cm}\left\| y^{(i)} - \left(x^{(i)}\right)^{'} \hspace{-.2cm} \hat \bw^{(\backslash k)}(\Lambda) \right\|^2\hspace{-.12cm}.\nonumber
\end{align}

To ease readability, without any loss of generality, in the rest of the paper we assume that all cross-validation folds have equal size, namely: $|\D_1| = \ldots = |\D_K| = N_V = \frac{N}{K}$. This implies that the amount of samples used in the validation step (see \eqref{eqn:lambdastar1}) of each cycle of the cross-validation is $\frac{N}{K}$, while $N_T = (K-1)\frac{N}{K}$    is the number of samples used to estimate the model parameters $ \hat \bw^{(\backslash k)}(\Lambda)$ in \eqref{eqn:thetaLS1}.

To leverage the matrix differential calculus approach, presented in Section \ref{Sec:approach}, and to  enhance notation clarity, we introduce the following matrix notation.
Let $X_{\backslash k} \in \mathbb{R}^{N_T \times \nw}$ and  $Y_{\backslash k} \in \mathbb{R}^{N_T \times \ny}$ be matrices stacking in their rows, respectively, the input samples  $x^{(i)}$ and the output samples $y^{(i)}$,  with $\left(x^{(i)}, y^{(i)}\right) \in \bigcup_{j \neq k} \D_{j} $. 
Similarly, let $X_{k} \in \mathbb{R}^{N_V \times \nw}$ and  $Y_{k} \in \mathbb{R}^{N_V \times \ny}$ be matrices stacking in their rows, respectively, the input  samples $x^{(i)}$ and the output samples $y^{(i)}$,  with $\left(x^{(i)}, y^{(i)}\right) \in  \D_{k} $.

Based on the introduced notation, the model parameter estimate \eqref{eqn:thetaLS1} can be rewritten in the following compact Tikhonov regularization form:
\begin{align} \label{eqn:thetaLS2}
    \hat \bw^{(\backslash k)}(\Lambda) &= \arg \min_{\bw } \, C(\bw, \Lambda) \nonumber \\ 
    &= \arg \min_{\bw} \, L(\bw) + R(\bw, \Lambda) \nonumber \\
    &=  \arg \min_{\bw} \, \frac{1}{2N_T}  \left\| Y_{\backslash k} - X_{\backslash k}\bw \right\|_F^2    + \frac{1}{2} \left\|  \Lambda \bw \right\|_F^2,
\end{align}

while the optimal hyperparameters  $\Lambda$ in \eqref{eqn:lambdastar1} can be expressed as:
\begin{align} \label{eqn:opt_Gamma_matrix}
\Lambda^{\star} &= \arg\min_{\Lambda } \, E\left(\hat \bw(\Lambda)\right) \nonumber \\ 
&= \arg\min_{\Lambda } \, \frac{1}{K}\sum_{k=1}^K L_k \left(\hat \bw^{(\backslash k)}(\Lambda)\right) \nonumber \\ 
&= \arg\min_{\Lambda } \, \frac{1}{K}\sum_{k=1}^K \frac{1}{2N_V} \left\| Y_{k} - X_{k} \hat \bw^{(\backslash k)}(\Lambda) \right\|_F^2.
\end{align}

\subsection{Motivation} \label{Sec:probform_motiv}

The main motivation behind the optimization of  multiple hyperparameters stems from the well-known fact that conventional regularization techniques, such as LASSO, Ridge regression, or Elastic Net, reduce the variance of the estimate while concurrently introducing bias in the estimation of model parameters, by shrinking them closer to zero. In a intuitive sense, employing multiple hyperparameters enables variance reduction of the estimate variance, while upholding an unbiased estimation of the model parameters.

In order to intuitively explain the concept, let us consider an illustrative example where the output variable $y$ is scalar and generated according to:
\begin{equation}
y = x'\bwo + e,
\end{equation}
where $e$ is white noise, independent of the input $x$, and $\bwo \in \mathbb{R}^{\nw}$ denotes the ``true'' (unknown) parameter vector. Let us take a scalar  $\bar{\lambda} \in \mathbb{R}$, $\bar{\lambda} \neq 0$, and let $\bwoj$ be the $j$-th element of $\bwo$. If the true parameter vector  $\bwo$ is assumed to be sparse, then a hyperparameter matrix $\Lambda\in \mathbb{R}^{D \times D}$ with diagonal elements
\begin{align} \label{eqn:lambda_opt}
\lambda_{j} = \left\{ \begin{array}{ll}
\bar{\lambda} & \textrm{ if } \bwoj = 0 \\
0 & \textrm{ if } \bwoj \neq 0
\end{array}
\right., \quad j=1,\ldots,D,
\end{align}
would regularize (thus  shrinking towards zero) the model parameters $\bwj$ such that the corresponding true parameters are  $\bwoj = 0$. On the other hand, it 
would not shrink towards zero  the model parameters $\bwj$  such that $\bwoj \neq  0$.  Hence, this type of regularization does not introduce bias into the model while it reduces the variance of the estimate compared to non-regularized approaches. Furthermore,  for $\bar{\lambda} \rightarrow \infty$, $\bwj \rightarrow 0$, and thus  input variable selection is also performed without introducing model bias.

Nevertheless, it is important to note, as also discussed in \cite{bengio2000gradient}, that using  multiple hyperparameters $\lambda$ introduces greater adaptability in the average loss $L_k\left(\hat \w^{(\backslash k)}(\lambda)\right)$ in \eqref{eqn:aver_loss}. More  flexibility could potentially increase  the risk of overfitting to the validation data employed for hyperparameter selection. In   Section \ref{sec:data_aug}, we discuss two strategies aimed at mitigating the risk of overfitting in the case where the true underlying parameter matrix $\bwo$ is sparse.

\section{Gradient-based multi-hyperparameter optimization}\label{Sec:approach}
In this section we show how to compute the gradient  $\nabla_{\Lambda} E$ of the evaluation criterion $E$ (appearing as a cost function of the outer optimization problem  \eqref{eqn:opt_Gamma_matrix})  with respect to the hyperparameter matrix $\Lambda$. This allows us to compute the optimal hyperparameters $\Lambda^\star$ through any gradient-based numerical optimization algorithm, such as Vanilla (stochastic) gradient descent;  ADAM, Adagrad; RMSProp; etc.

The estimation of the model parameters $\hat \bw^{(\backslash k)}(\Lambda)$ in the Tikhonov regularized form \eqref{eqn:thetaLS2} has the following analytical solution: 
\begin{align} \label{eqn:tikhonov_solution}
\hat \bw^{(\backslash k)}(\Lambda) = \left(X_{\backslash k}'X_{\backslash k} + N_T\Lambda\Lambda\right)^{-1}X_{\backslash k}'Y_{\backslash k},
\end{align}
where the transpose of $\Lambda$ is omitted  in the equation above since  $\Lambda$ is diagonal, thus  symmetric.

The gradient $\nabla_{\Lambda} E$ can be readily computed via automatic differentiation  (e.g., using the deep learning Python libraries such as  PyTorch or TensorFlow). Nevertheless, employing backpropagation to solve the Tikhonov regularization problem \eqref{eqn:thetaLS2} was  observed to be inefficient when handling a growing number of features, as also discussed in the numerical example in  Section \ref{Sec: example_perf}. For this reason, an analytical strategy for gradient computation is adopted in this paper. The proposed approach leverages the framework of matrix differential calculus \cite{magnus2019matrix}, enabling  the derivation of gradients in a matrix form, thereby exploiting the efficient matrix-vector multiplication features available in scientific computing libraries such as NumPy, PyTorch, and JAX.

\begin{subequations}
The following expression 
for the gradient $\nabla_{\Lambda} E$ is derived:
\begin{align}
\nabla_{\Lambda} E &= - \frac{N_T}{K N_V} \sum_{k=1}^K \mathrm{vec}\left(\Lambda B_k + B_k \Lambda \right), \label{eqn:gradient_eval_criterion_Lambda}
\end{align}
with $B_k=A_k' X_k' R_k \left(\hat \bw^{(\backslash k)}\right)'$,   $A_k=\left(X_{\backslash k}'X_{\backslash k} + N_T\Lambda\Lambda\right)^{-1}$, and  $R_k=X_{k} \hat \bw^{(\backslash k)}(\Lambda) - Y_{k}$. 

Since $\Lambda$ is constrained to be diagonal, at each update step of gradient-descent, only the gradient of the evaluation criterion  $E$ with respect to the diagonal elements $\lambda$ of the hyperparameter matrix $\Lambda$  is of practical interest, and its expression is given by:
\begin{align}
\nabla_{\lambda} E &= - \frac{N_T}{K N_V} \sum_{k=1}^K \mathrm{diag}\left(\Lambda B_k + B_k \Lambda \right).  \label{eqn:gradient_eval_criterion_lambda}
\end{align}

All the details behind the derivation of the gradients \eqref{eqn:gradient_eval_criterion_Lambda} and \eqref{eqn:gradient_eval_criterion_lambda}  are reported in Appendix \ref{app-der1}. 

\end{subequations}

\section{Reducing the risk of overfitting} \label{sec:data_aug}

Assuming that the true parameter matrix $\bwo$ is sparse, this section introduces two strategies with the objective of guiding the hyperparameters $\Lambda$ towards the form outlined in \eqref{eqn:lambda_opt}. These strategies aim to mitigate the potential of overfitting the hyperparameters $\Lambda$ during the cross-validation stage.

\subsection{Optimal-solution augmentation}
\label{Sec:opt_sol_data}

The strategy discussed in this subsection is based on the observation that any non-zero scaling of the hyperparameter matrix $\Lambda$ does not affect its ``optimal'' structure  in \eqref{eqn:lambda_opt}. Thus, given: i) true model parameters $\bwo$, and ii) the optimal structure of $\Lambda^\star$ from \eqref{eqn:lambda_opt}, i) and ii) implies that the   term $\left\| \gamma \Lambda^\star \bwo \right\|_F^2$,  equals $0$ for any scaling parameter $\gamma \in \mathbb{R}$ ($\gamma \neq 0$). Nevertheless, the model parameters $\bwj$ associated with zero elements of the true parameters  $\bwoj$ (or equivalently associated with non-zero hyperparameter $\lambda^\star_j$)   remain subject to regularization towards  zero.

To promote the sparse structure of $\Lambda^\star$ in \eqref{eqn:lambda_opt}, we consider a finite set $\Gamma$ of non-zero scaling parameters $\gamma$. The optimal hyperparameters $\Lambda^{\star}$ are then computed as follows:
\begin{align} \label{eqn:opt_Gamma_matrix_gamma}
\Lambda^{\star} = \arg\min_{\Lambda} \frac{1}{2N} \frac{1}{|\Gamma|} \sum_{k=1}^K \sum_{\gamma \in \Gamma} \left\| Y_{k} - X_{k} \hat \bw^{(\backslash k)}(\gamma \Lambda) \right\|_F^2,
\end{align}
with model parameters given by:
\begin{align} \label{eqn:thetaLS1_gamma}
&\hat \bw^{(\backslash k)}(\gamma\Lambda) = \\&\arg \min_{\bw} \frac{1}{2N_T} \sum\limits_{\substack{(x^{(i)},y^{(i)}) \\\in \bigcup_{j \neq k} \D_{j} }} \left\| y^{(i)} - \left(x^{(i)}\right)' \bw \right\|^2 + \frac{1}{2} \left\| \gamma\Lambda \bw \right\|_F^2.\nonumber
\end{align}

This approach bears similarity to data augmentation, a well-established technique often employed in deep learning tasks such as image recognition. Data augmentation involves introducing synthetic images (generated through operations like, for instance, rotation or adjustments in illumination) to the training dataset. This helps enforce desirable properties such as invariance to rotations and illumination. Instead of adding artificial data, our approach ensures that the hyperparameters $\Lambda$  are not  just adjusted to match the data, but rather ensure a small fitting error $\left\| Y_{k} - X_{k} \hat \bw^{(\backslash k)}(\gamma \Lambda) \right\|_F^2$ in \eqref{eqn:opt_Gamma_matrix_gamma}, for any $\gamma \in \Gamma$, as anticipated by the theoretical optimal hyperparameters $\Lambda^\star$ in \eqref{eqn:lambda_opt}. In other words, the optimal model parameters $\bw^{(\backslash k)}(\gamma \Lambda)$ are expected to be invariant to any non-zero scaling factor $\gamma$. To further highlight the similarity with data augmentation, it is interesting to note that  newly constructed artificial model parameters $\hat \bw^{(\backslash k)}(\gamma\Lambda)$ (for each $\gamma \in \Gamma$) are used in the computation of $\Lambda$ in \eqref{eqn:opt_Gamma_matrix_gamma}.

The practical implementation of the regularization strategy discussed above requires minor changes to \eqref{eqn:tikhonov_solution}, \eqref{eqn:gradient_eval_criterion_Lambda} and \eqref{eqn:gradient_eval_criterion_lambda}, which become, respectively:

\begin{align} 
&\hat \bw^{(\backslash k)}(\gamma\Lambda) = \left(X_{\backslash k}'X_{\backslash k} + N_T\gamma^2\Lambda\Lambda\right)^{-1}X_{\backslash k}'Y_{\backslash k} \,, \label{eqn:tikhonov_solution_gamma} \\
&\nabla_{\Lambda} E = - \gamma^2\frac{N_T}{K N_V} \frac{1}{|\Gamma|} \sum_{k=1}^K \sum_{\gamma \in \Gamma}\mathrm{vec}\left(\Lambda \tilde{B}_k + \tilde{B}_k \Lambda \right) \,, \label{eqn:gradient_eval_criterion_Lambda_gamma} \\
&\nabla_{\lambda} E = - \gamma^2 \frac{N_T}{K N_V} \frac{1}{|\Gamma|} \sum_{k=1}^K \sum_{\gamma \in \Gamma}\mathrm{diag}\left(\Lambda \tilde{B}_k + \tilde{B}_k \Lambda \right) \,. \label{eqn:gradient_eval_criterion_lambda_gamma}
\end{align}

with $\tilde{B}_k=\tilde{A}_k' X_k' R_k \left(\hat \bw^{(\backslash k)}\right)'$,   $\tilde{A}_k=\left(X_{\backslash k}'X_{\backslash k} + N_T\gamma^2\Lambda\Lambda\right)^{-1}$, and  $R_k=X_{k} \hat \bw^{(\backslash k)}(\Lambda) - Y_{k}$.

\subsection{Regularization in validation}
The second strategy  for reducing the risk of overfitting stems from the same considerations discussed in the previous paragraph. Indeed, in the case of a sparse true parameter vector $\bwo$, we would aim to guide the hyperparameter $\Lambda$ to take the structure  in \eqref{eqn:lambda_opt}, rather than simply minimizing the fitting herror $\left\| Y_{k} - X_{k} \hat \bw^{(\backslash k)}(\Lambda) \right\|_F^2$ of the $k$-th validation fold.  To this aim,   the validation loss function in \eqref{eqn:opt_Gamma_matrix} is modified as follow, by introducing a regularization term:
\begin{align} \label{eqn:opt_Gamma_matrix_reg}
\Lambda^{\star} =   \arg\min_{\Lambda} & \frac{1}{2N}\sum_{k=1}^K   \left\| Y_{k} - X_{k} \hat \bw^{(\backslash k)}(\Lambda) \right\|_F^2 +  \nonumber \\
 & +  \frac{1}{2} \mu \sum_{k=1}^K\left\|  \Lambda \hat \bw^{(\backslash k)}(\Lambda) \right\|_F^2, 
\end{align}
 where $\mu \in \mathbb{R}$, with $\mu > 0$, acts as an additional hyperparameter. Thus, unlike the ``optimal-solution augmentation'' strategy discussed in Section \ref{Sec:opt_sol_data}, this approach requires the tuning of an additional (albeit scalar) regularization hyperparameter, hence the requirement for an additional hyper-validation dataset.

The practical implementation of the regularization strategy discussed in this subsection requires the derivation of the gradient of the regularization term:
\begin{align} \label{eqn:Q}
\Q = \frac{1}{2} \mu \sum_{k=1}^K\left\|  \Lambda \hat \bw^{(\backslash k)}(\Lambda) \right\|_F^2,
\end{align}
with respect to $\Lambda$ (or equivalently $\lambda$), whose analytical expressions are given by:

\begin{subequations}

\begin{align}
\nabla_{\Lambda}\Q &= \mu \sum_{k=1}^K \mathrm{vec}\left( D_k \left(\hat \bw^{(\backslash k)}\right)' - N_T \left(\Lambda G_k + G_k \Lambda \right)\right), \label{eq:grad_reg_criterion_final_Lambda}\\ 
\nabla_{\lambda}\Q &= \mu \sum_{k=1}^K \mathrm{diag} \left( D_k \left(\hat \bw^{(\backslash k)}\right)' - N_T \left(\Lambda G_k + G_k \Lambda \right) \right), \label{eq:grad_reg_criterion_final_lambda}
\end{align}
with $D_k = \Lambda \hat \bw^{(\backslash k)}$ and $G_k=A_k'\Lambda'D_k \left(\hat \bw^{(\backslash k)}\right)'$.

Derivations of the gradients $\nabla_{\Lambda}\Q$ (resp. $\nabla_{\lambda}\Q$) in \eqref{eq:grad_reg_criterion_final_Lambda} (resp. \eqref{eq:grad_reg_criterion_final_lambda}) are based on similar differential matrix calculus arguments used to compute \eqref{eqn:gradient_eval_criterion_Lambda} and \eqref{eqn:gradient_eval_criterion_lambda}, and  are reported in Appendix \ref{app-der2}.

\end{subequations}

\section{Examples} \label{Sec: example}

In this section, we present numerical examples to demonstrate the effectiveness of the multi-hyperparameter optimization approach. In this context, we conduct a comparative analysis between this  approach and established benchmark techniques, such as LASSO, Ridge  and Elastic Net regressions, all of which are readily available through the Python \emph{scikit-learn} package. Comparison   in terms of final model performance and computational complexity is discussed in Section \ref{Sec:exampleFeatures} and \ref{Sec: example_perf}, respectively. Parametric identification of \emph{Linear Parameter-Varying} (LPV) dynamical systems with an unknown model structure is addressed  in Section~\ref{Sec:exampleLPV}.

Model performance is measured on a test set (used neither for training nor for validation) through the R$^2$ index, defined as:
\begin{align} \label{eqn:r2}
\textrm{R}^2 = \max \left\{ 0; 1 - \frac{\sum_{i=1}^{N_{\rm test}} \left\|   y^{(i)} - \hat{y}^{(i)} \right\|^2 }{\sum_{i=1}^{N_{\rm test}} \left\|   y^{(i)} - \bar{y} \right\|^2}\right\},
\end{align}
where  $\hat{y}^{(i)}$ is the estimated model output for the $i$-th sample, $\bar{y}$ is the sample mean of the outputs over the test dataset, and $N_{\rm test}$ is length of the test dataset.  In assessing  algorithms' performance, we use  test datasets  larger (up 100x) than the training set. Although this is not realistic in practice, it allows us to have a fair and statistically significant evaluation and comparison of different algorithms.

Output data (including test data) is corrupted by an additive noise $e$. The effect of the noise is measured in terms of  the Signal-to-Noise Ratio (SNR), defined as:
\begin{equation}\label{eqn:SNR}
\mathrm{SNR}=10\log{\left( \frac{\left\|y-e\right\|^2}{\left\|e\right\|^2} \right)}.
\end{equation}

For the multi-ridge case  discussed in the paper, numerical optimization  is performed through  gradient descent, with learning rate  chosen manually. In the first and third examples and in all   algorithms discussed in this section, 5-fold cross-validation is implemented to select the regularization hyperparameters. Once the hyperparameters are chosen,  models are re-estimated  on the complete training dataset. In the second example, a simple holdout validation is used, as the focus is only on the numerical computation of the gradient $\nabla_{\lambda} E$ and not on assessing model performance.

All computations are performed on a server with two 64-core AMD EPYC 7742 Processors, 256 GB of
RAM, and 4 Nvidia RTX 3090 GPUs. The hardware resources of the server have been limited to 10 CPU threads and 1 GPU.

The software has been developed in PyTorch with a scikit-learn compatible interface. All the codebase, along with documentation, is available in the GitHub repository \href{https://github.com/gabribg88/Multiridge}{https://github.com/gabribg88/Multiridge}.

\subsection{Example 1: Sensitivity to the number of features}  \label{Sec:exampleFeatures}
In this numerical example, we investigate the sensitivity of different  regularization strategies with respect to the number of features,  while keeping the number of data points and the information ratio (i.e., the ratio between the total number of input features available in the dataset  and the number of informative features) constant. 

Since the number of features effectively determines the complexity of the model, we expect that as the ratio of input  features to data points grows, the model's capacity to capture not only the underlying pattern but also the noise within the training data, increases, potentially resulting in overfitting. As the overfitting increases, the impact of the regularization term becomes more significant. This provides a basis for comparing the effectiveness of various regularization strategies.

$14$ distinct conditions are simulated, with the number of features $D$ incrementally ranging from $100$ to $1400$ with a step of $100$. To enhance the statistical significance of the results,  we replicated each condition $10$ times, utilizing different random seeds for each replication. 
In each individual experiment, a total of $N_{train}=10^3$ training data samples and $N_{test}=10^5$ test samples were generated, according to the following linear data generation procedure:

\begin{align}
    y = x'\tilde{\bw}_{\rm o} + e,
\end{align}
where each component of the feature vector $x \in \mathbb{R}^{\nw}$ is drawn from a Normal distribution $ \mathcal{N}(0, 1)$. The noise $e$ is also Normally distributed, with $e \sim \mathcal{N}(0, \sigma_e^2)$ and variance $\sigma_e^2$ chosen to ensure an SNR of approximately $20$ dB.
The actual coefficient vector $\tilde{\bw}_{\rm o}$ is a sparse version of a vector $\bw_{\rm o}$, whose components are drawn from a uniform distribution $\mathcal{U}_{\,[-50, 50]}$ between $-50$ and $50$. The relation between $\tilde{\bw}_{\rm o}$ and $\bw_{\rm o}$ is thus expressed as:
\begin{align} \label{eqn:theta_sparse}
\tilde{\bw}_{{\rm{o}},j} = \left\{ \begin{array}{ll}
\bw_{{\rm{o}},j} & \textrm{ if } j \in \mathcal{I} \\
0 & \textrm{ if } j \notin \mathcal{I}
\end{array}
\right., 
\end{align}
where $\mathcal{I} \subseteq \{1,2,\dots, D\}$ represents the index set of informative features. The cardinality of  $\mathcal{I}$ is $|\mathcal{I}| = \left\lfloor D I_{frac}\right\rfloor$, with $\left\lfloor  \cdot \right\rfloor$ being the floor operator.   The  fraction of informative features $I_{frac}$ is set to $0.5$ and kept constant throughout this example.  Each element $j \in \mathcal{I}$ is uniformly drawn  at random, without replacement,  in $\{1,2,\dots, D\}$.

In each experiment, the benchmark learning algorithms underwent a systematic  training process with the following sequential stages: i) standardization of both features and target variable; ii) hyperparameter optimization  through  5-fold cross-validation;, iii) re-training of the models on the complete training dataset;  iv) evaluation of algorithms' performances on a test set. Hyperparameter optimization  varies depending on the regularization algorithm, as detailed below:
\begin{itemize}
    \item Ridge regression: a grid-search was conducted on a logarithmically spaced grid of $10^3$ values for the $l_2$-regularization strength. The grid values ranged from $10^{-3}$ to $10^6$.
    \item LASSO: a grid-search was performed using a logarithmically spaced grid of $10^3$ values for the $l_1$-regularization strength. The grid values spanned from $10^{-5}$ to $10^2$.
    \item Elastic Net regression: a random-search was executed with $10^3$ repetitions on pairs of: logarithmically spaced values of the regularization strength ranging from $10^{-5}$ to $10^3$,  and  ratio between $\ell_1$ and $\ell_2$ regularization spanning linearly from $0$ to $1$.
\end{itemize}

For multi-ridge regression,  the hyperparameters $\Lambda$ are optimized through a gradient-based approach, and initialized as an identity matrix $I_D$. The learning rate is  set to  $350$ across all scenarios, with a decay rate of $0.999$ applied at each epoch over the course of $300$ training epochs. %

\begin{figure}[!tb]
\centering
\includegraphics[width=0.48\textwidth]{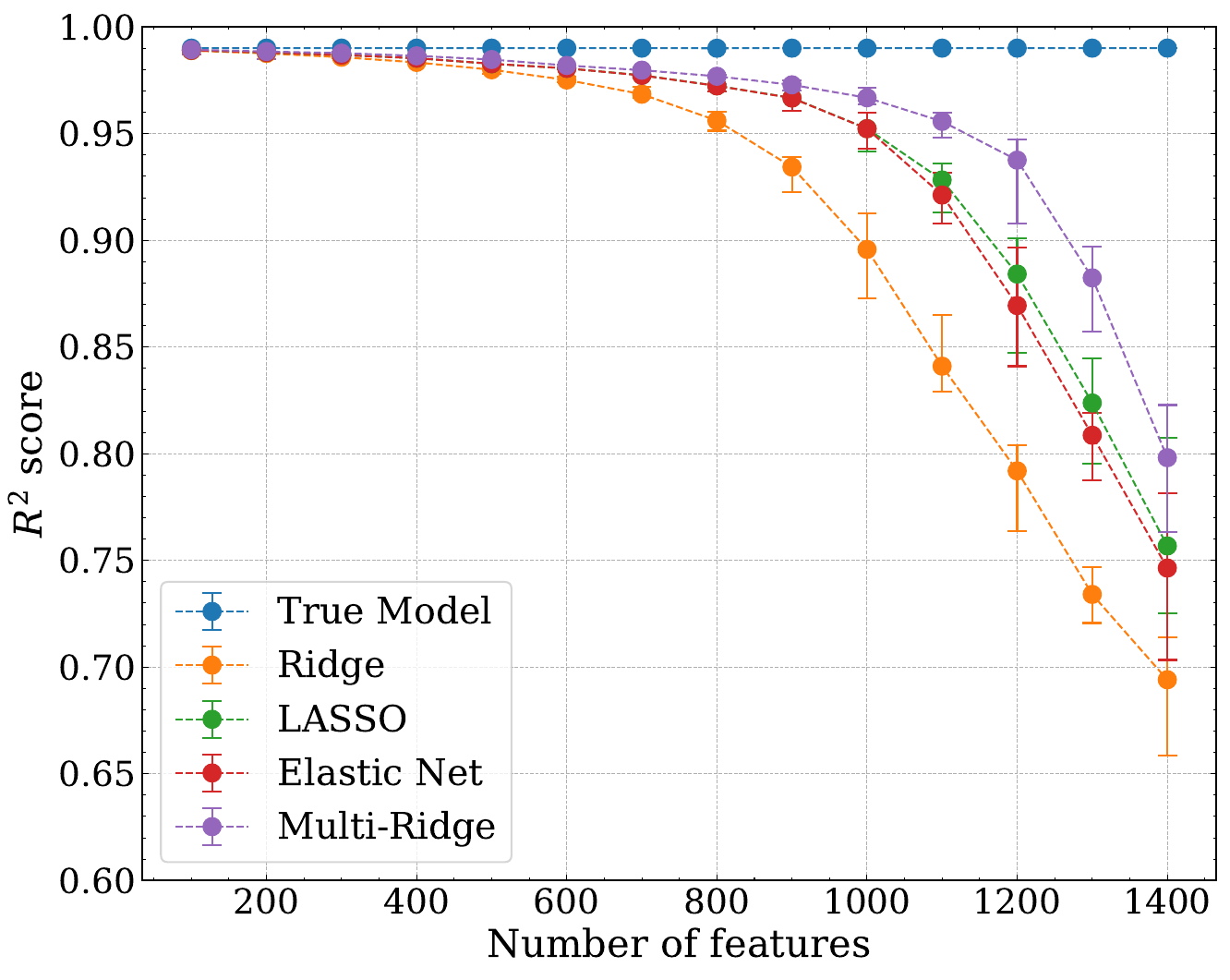}
\caption{Example 1:  model performance  \emph{vs} number of input features for different regularization algorithms. Median over $10$ repetitions (dots), along with $5$-th and $95$-th percentile (bars) are depicted.}
\label{fig:Massive_experiment}
\end{figure}

Results are presented in Fig.~\ref{fig:Massive_experiment}, illustrating the sensitivity of model performance with respect to the number of input features for Ridge, LASSO, Elastic Net, and the proposed  multi-ridge  regression. For   less than $300$  features, all methods deliver performance comparable to the oracle model with the true parameter vector $\tilde{\bw}_{\rm o}$. However, as the number of input features grows, each method exhibits a gradual decline in performance due to overfitting. 

Multi-ridge regression  consistently outperforms the other benchmark regularization strategies. Notably, simple Ridge regression records the lowest performance. This outcome is expected for the given problem with a sparse true parameter vector. Thus, algorithms that promote sparse solutions, such as LASSO, are more suitable. Nevertheless, even though multi-ridge regression  employs $\ell_2$ regularization instead of $\ell_1$,  it still outperforms LASSO.  This enhanced performance stems from our method's flexibility in assigning a distinct regularization strength to each feature, based on its inherent informativeness.

\subsection{Example 2: Computational performances} \label{Sec: example_perf}

In this numerical example we show the advantages in terms of both computational efficiency (measured in terms of GPU time) and numerical accuracy (assessed by quantifying the norm of the difference between the analytic calculation and the automatic differentiation, as implemented within the PyTorch framework) during the computation of the gradient in \eqref{eqn:gradient_eval_criterion_lambda}. These benefits become increasingly pronounced as the number of features in the simulated dataset grows. More specifically,  we constructed  $50$ distinct scenarios, each characterized by a progressively growing number of features, with logarithmically spaced values. For every scenario, we conducted $10$ experiments, each repeated with varying random seeds to enhance the statistical significance of the results.

In all the experiments, a fixed number of $N_{train}=10^3$ training data samples were generated. Each data sample was composed of an input-output pair $(x,y)$, where $x \in \mathbb{R}^D$ and $y \in \mathbb{R}^M$, with $M=10$. Generated samples were subsequently partitioned into training fold data and validation fold data, according to an 80:20 ratio. This is equivalent to simple holdout validation instead of $K$ fold.

The input features $x$, the targets $y$, as well as the hyperparameters $\lambda$, are generated according to a Normal distribution, with zero mean and unitary variance.

\begin{figure*}
\begin{subfigure}[h]{0.48\linewidth}
\includegraphics[width=\linewidth]{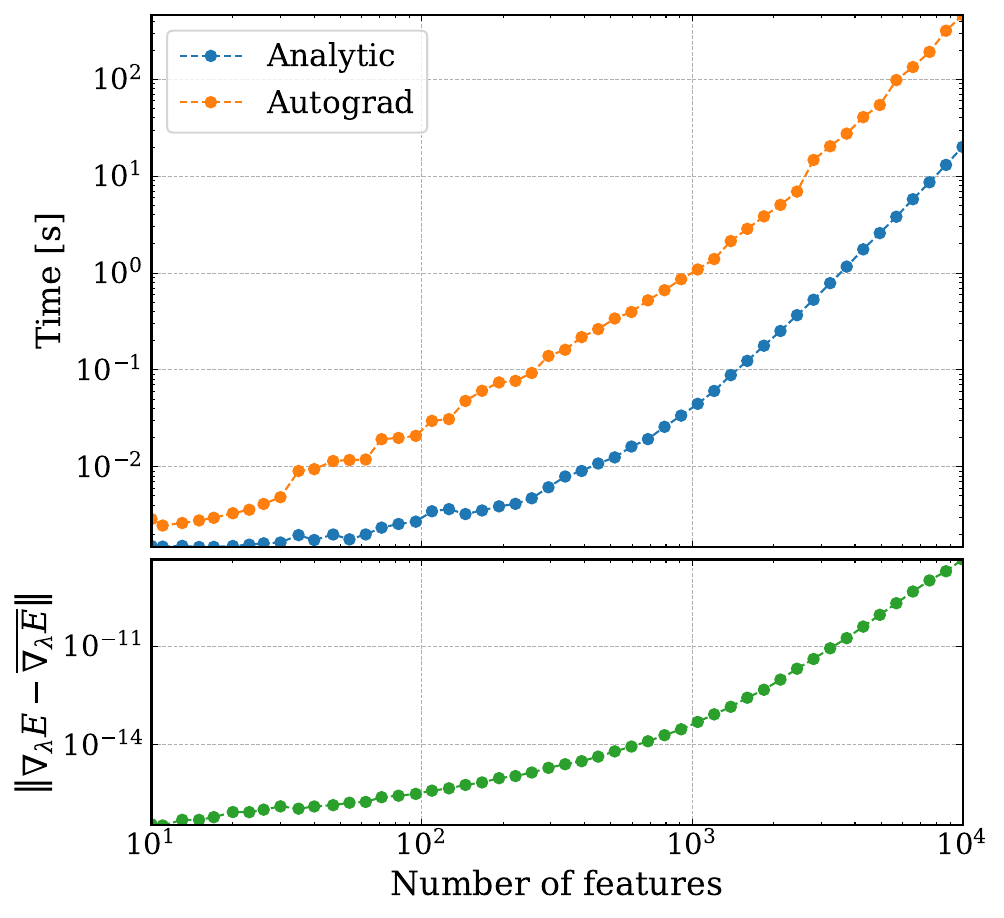}
\caption{Double-precision floating-point format (float64)}
\end{subfigure}
\hfill
\begin{subfigure}[h]{0.48\linewidth}
\includegraphics[width=\linewidth]{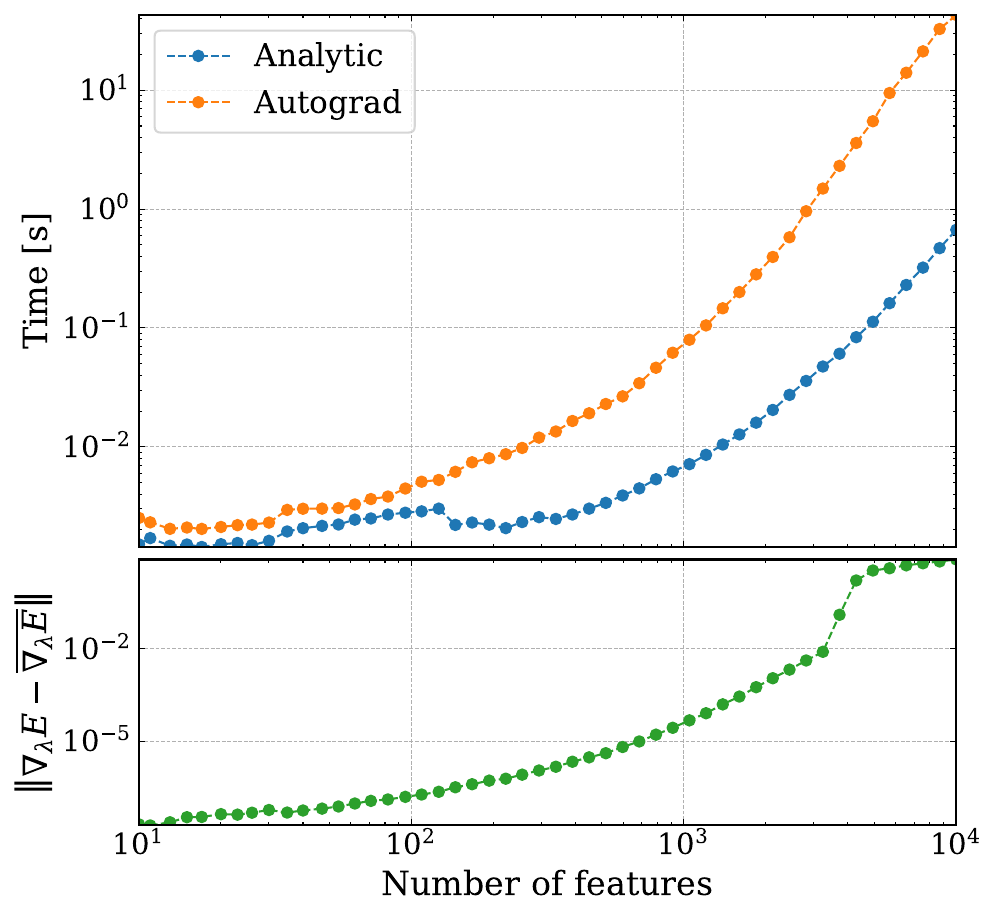}
\caption{Single-precision floating-point format (float32)}
\end{subfigure}%
\caption{Example 2: Results of computational performance. GPU time required to compute the gradient $\nabla_{\lambda}E$ (upper panels); norm of the difference between analytic computation of the gradient $\nabla_{\lambda}E$ and its numerical computation through automatic differentiation in PyTorch, denoted  as $\overline{\nabla_{\lambda}E}$ (lower panels).}
\label{fig:Computational_performances}
\end{figure*}

The experiment is conducted in 2 different scenarios. Firstly, with variables represented in double-precision floating-point format (float64), and secondly, with variables represented in single-precision floating-point format (float32). The number of features ranges from $10$ to $10^4$.

Results are illustrated in Fig.~\ref{fig:Computational_performances}. The upper panel of the figure displays the GPU time required for the computation of the gradient $\overline{\nabla_{\lambda}E}$. The orange line corresponds  to the time taken by the PyTorch autograd, whereas the blue line indicates the time consumed by the analytic calculation. As the number of features increases, the analytic calculation showcases a significant enhancement in computational efficiency, obtaining computational speed more than an order of magnitude faster w.r.t. automatic differentiation.

The lower panel of the figure displays the norm of the difference between the analytic calculation and the computation through automatic differentiation. In the scenario where variables are represented in double-precision floating-point format, this difference is negligible. Nonetheless, when variables are represented in single-precision floating-point format, this difference becomes noteworthy as the number of features increases.

\subsection{LPV identification} \label{Sec:exampleLPV}

As a last example, we consider the identification of an LPV system described by the AutoRegressive eXogenous (ARX) form:
\begin{align} \label{eqn:systemLPV}
y(k)  = &\;0.5\cos{(p(k))}y(k-2) -0.1 \sin^2{(p(k))}y(k-3) \nonumber  \\
& + (\cos{(p(k))}-\sin{(p(k))})u(k-2) \nonumber\\&+3\sin{(p(k))}u(k-3)  
 + e(k),
\end{align}
where $u(k)$, $y(k)$, and $p(k)$ represent  the measurement of the input, output, and scheduling variable at time step $k$, respectively, while $e(k)$ is a random Gaussian white noise.

A small-size dataset $\D$ of size $N=50$ is gathered by exciting the system \eqref{eqn:systemLPV} with a  white noise input sequence $\{u(k)\}_{k=1}^N$, with $u(k)$ drawn from a Gaussian distribution $\mathcal{N}(0, 1)$. The scheduling variable $p(k)$ is also drawn from a Gaussian distribution, with $p(k) \sim \mathcal{N}(0, \pi)$. The  amplitude of the noise $e(k)$ is chosen such that the SNR 
is about $14$ dB (namely, the ratio between  the power of the noise and the power of the noise-free output  is approximately $4$\%).

In estimating the LPV system \eqref{eqn:systemLPV}, we assume to be in a setting where the system structure is not known, and we consider an over-parameterized LPV-ARX model of the form:
\begin{align} \label{eqn:modelLPV}
\hat{y}(k)=\sum_{j=1}^{\na}{a}_j(p(k))y(k-j)+\sum_{j=1}^{\nb}{b}_{j}(p(k))u(k-j), \nonumber
\end{align} 
with $\na=\nb=30$. Furthermore, the dependence of the coefficient functions ${a}_j$ and ${b}_j$ ($j=1,\ldots,30$) on   the scheduling parameter $p$ is  parameterized in terms of a  linear combination of a finite number of \emph{a-priori}   chosen basis functions, as follows: 
\begin{subequations} \label{eq:abpar}
\begin{align}
a_j(p(k))= & \sum_{s=1}^{n_{\alpha}}a_{j,s}\psi_{s}(p(k)), &  i=1,\ldots,\na, \\
b_j^{\mathrm{}}(p(k))= & \sum_{s=1}^{n_{\alpha}}b_{j,s}\psi_{s}(p(k)), &  j=1,\ldots,\nb,
\end{align}
\end{subequations}
where $n_\alpha = 8$, $\{a_{j,s} \in \mathbb{R}\}_{s=1}^{n_\alpha}$, and $\{b_{j,s} \in \mathbb{R}\}_{s=1}^{n_\alpha}$  are the  unknown constant parameters to be estimated,  and $\psi_{s}(p(k))$ is the $s$-th element of the vector function:
\begin{align}
    \psi_{s}(p) = \left[1 \ p \ p^2 \ p^3 \ \sin(p) \ \cos(p) \ \sin^2(p) \ \cos^2(p) \right].
\end{align}

Summarizing, we have to estimate $n_\alpha  (\na + \nb) = 480$  parameters   from a dataset of length $N=50$. In such a setting, regularization is expected to play a fundamental role for the estimation process.

 For multi-ridge regression, the  initial conditions of the hyperparameters $\lambda$ are extracted from the solution of a LASSO problem. More specifically, if the model parameter $\bw_{i}$ estimated by the LASSO is  zero, then the related hyperparameter $\lambda_j$ is set to $10$, otherwise $\lambda_j=1$. This means that, in the initialization of $\lambda$, a 10x higher penalization is given to the model parameters estimated as zero by the LASSO. The optimal-solution augmentation strategy discussed in Section~\ref{Sec:opt_sol_data} is implemented to reduce the risk of overfitting to the validation datasets, by minimizing the  evaluation  criterion in \eqref{eqn:opt_Gamma_matrix_gamma} for $\gamma \in \Gamma = \{0.5, \ 1, \ 2\}$.
 
\begin{figure}[!tb]
\centering
\includegraphics[width=0.48\textwidth]{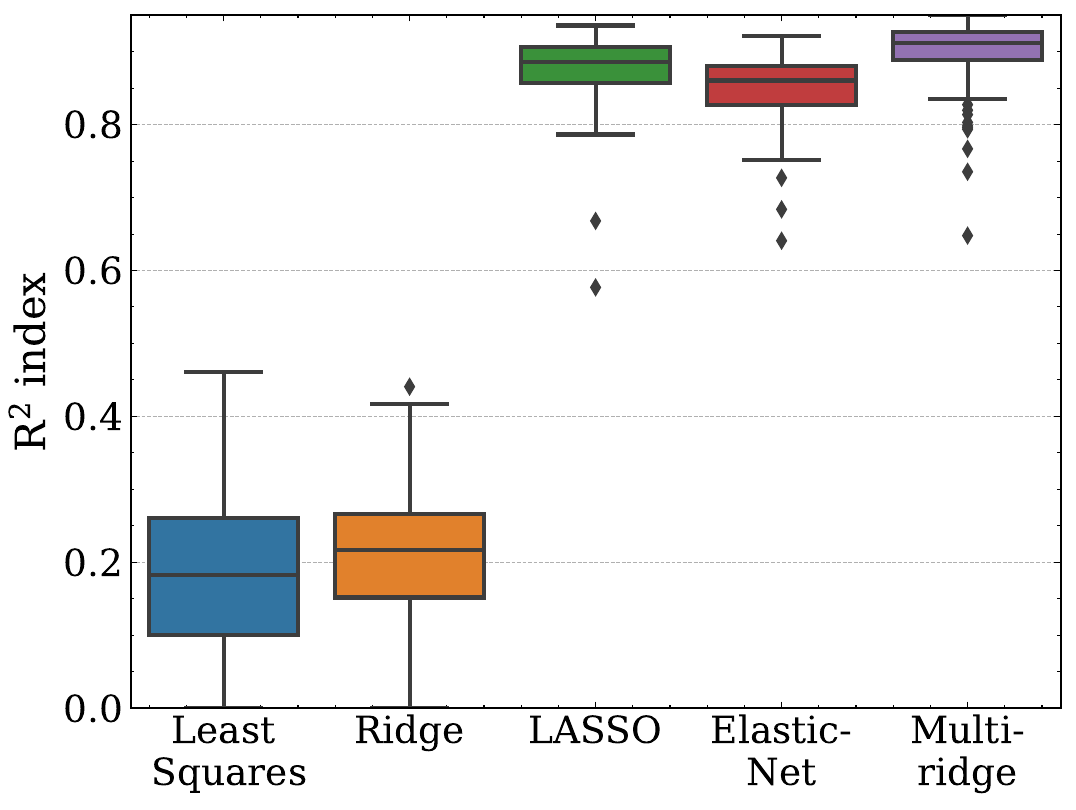}
\caption{LPV identification: boxplots of the R$^2$ index over 200 Monte Carlo runs. Medians: Least Squares (0.18); Ridge (0.21); LASSO (0.88); Elastic Net (0.86); multi-ridge (0.91). }
\label{fig:LPV_BP}
\end{figure}

The performance of all the tested algorithms is evaluated on a   test data of length $N_{\rm test}=3000$. Figure \ref{fig:LPV_BP} shows the boxplots of the different approaches over $200$ Monte Carlo simulations, with: input, noise and scheduling variable independently regenerated at each run. The figure shows that LASSO, Elastic Net and the  multi-hyperparameter approach proposed  in this paper provide satisfactory performance, achieving a median R$^2$ value of $0.88$, $0.86$ and $0.91$, respectively. However, this example also highlights how simple Ridge regression does not have enough flexibility in its regularization term. In contrast,  multi-ridge regression, having a much larger flexibility, surpasses LASSO's performances too, even when dealing with a sparse structure in the  ``true'' parameter vector.

\section{Conclusions} \label{Sec:conclusion}
Multi-penalty Ridge   regression  increases the degrees of freedom of the regression algorithm, thus allowing to enhance   model's performance compared to widely adopted regularization techniques such as LASSO, Ridge, and Elastic Net regression. The optimization of multiple regularization hyperparameters can be achieved through gradient descent. Here, the analytical expression of the gradient for a cross-validation criterion with respect to the regularization hyperparameters can be derived using matrix differential calculus, avoiding  the need for automatic differentiation and backpropagation. As shown in the numerical examples, this offers numerical benefits, especially when dealing with a large number of features.

The approach is specifically tailored to $\ell_2$ regularization and takes advantage of the fact that the optimal model parameters can be determined analytically. However, ongoing research aims to extend the matrix differential calculus-based approach presented in this  paper to other types of problems where  optimal model parameters cannot be expressed analytically and must be determined through iterative numerical optimization. These types of problems include  LASSO with multiple hyperparameters or  nested optimization problems arising in meta-learning, where models are trained  on the task of learning themselves, often with the aim of rapidly adapting to new tasks using only a small-size dataset.

\bibliographystyle{plain}
\bibliography{Bibliography.bib}

\section*{Appendix}
\appendix
\section{Derivation of \eqref{eqn:gradient_eval_criterion_Lambda} and \eqref{eqn:gradient_eval_criterion_lambda}} \label{app-der1}
By the \emph{first identification theorem}  \cite[Chapter 9, Section 5]{magnus2019matrix},  the relationship between the differential and the gradient of the evaluation criterion $E$ is given by:
\begin{subequations} \label{eqn:dE_app}
\begin{align} \label{eqn:first_identification}
dE & = \left(\nabla_{\Lambda} E\right)' d\mathrm{vec}\Lambda,
\end{align}
or equivalently:
\begin{align} \label{eqn:first_identification_lambda}
dE & = \left(\nabla_{\lambda} E\right)' d\lambda. 
\end{align}
\end{subequations}
Thus, from \eqref{eqn:dE_app}, both   $\nabla_{\Lambda} E$ and $\nabla_{\lambda} E$  can be readily obtained, as discussed in the following.
By differentiating both sides of \eqref{eqn:eval_criterion} and leveraging linearity of the differential operator, we obtain:  
\begin{align} \label{eqn:diff_eval_criterion}
dE = d\left(\frac{1}{K}\sum_{k=1}^K L_k\right) =  \frac{1}{K}\sum_{k=1}^K dL_k. 
\end{align}
To derive $dL_k$ (and thus  $dE$),    $L_k$ is rewritten in terms of the trace operator as follows:  
\begin{align} \label{eqn:eval_criterion2}
L_k &= 
 \frac{1}{2N_V} \left\| R_k \right\|_F^2 
= \frac{1}{2N_V} \mathrm{tr}\left(R_k'R_k\right) \,,
\end{align}
where $R_k=X_{k} \hat \bw^{(\backslash k)}(\Lambda) - Y_{k}$.
By differentiating \eqref{eqn:eval_criterion2}:
\begin{align} \label{eqn:diff_eval_criterion2}
dL_k &= \frac{1}{2N_V} d\mathrm{tr}\left(R_k'R_k\right) \nonumber \\
&= \frac{1}{2N_V} \mathrm{tr}\left[\left(dR_k'\right)R_k + R_k'dR_k\right] \nonumber \\
&= \frac{1}{2N_V} \mathrm{tr}\left[\left(dR_k\right)'R_k + R_k'dR_k\right] \nonumber \\
&= \frac{1}{2N_V} \mathrm{tr}\left(R_k'dR_k + R_k'dR_k\right) \nonumber \\
&= \frac{1}{N_V} \mathrm{tr}\left(R_k'dR_k\right), \nonumber \\
\end{align}
where the following properties of the differential operator are used, for any matrix  $U \in \mathbb{R}^{m \times n}$ and  $V \in \mathbb{R}^{m \times p}$: 
\begin{subequations} \label{eqn:usef_eqn}
\begin{align}
&d(U') = dU', \\
&d\left(UV\right) = \left(dU\right)V + UdV  \quad \textit{(Product rule)}. \label{eqn:product_rule}
\end{align}
\end{subequations} 
The computation of $dL_k$ in \eqref{eqn:diff_eval_criterion2} requires to compute the differential $dR_k$. From the definition of $R_k$ and the analytical expression of the model parameters $\hat \bw$ in \eqref{eqn:tikhonov_solution}: 
\begin{align} \label{eqn:dR}
d R_k  & = d\left(X_{k} \hat \bw^{(\backslash k)}\right)  \\
 & = d\left(X_{k} \left(X_{\backslash k}'X_{\backslash k} + N_T\Lambda\Lambda\right)^{-1}X_{\backslash k}'Y_{\backslash k} \right) \nonumber \\
& = d\left(X_{k} A_k X_{\backslash k}'Y_{\backslash k} \right) = R_k'X_{k} \left(dA_k\right) X_{\backslash k}'Y_{\backslash k},\nonumber
\end{align}
with $A_k = \left(X_{\backslash k}'X_{\backslash k} + N_T\Lambda\Lambda\right)^{-1}$.
By substituting \eqref{eqn:dR} into \eqref{eqn:diff_eval_criterion2}, and from the cyclic property of the trace operator,  we obtain: 
\begin{align} \label{eqn:dLk}
dL_k  
&= \frac{1}{N_V} \mathrm{tr}\left[R_k'X_{k} \left(dA_k\right) X_{\backslash k}'Y_{\backslash k}\right] \nonumber \\
&= \frac{1}{N_V} \mathrm{tr}\left(X_{\backslash k}'Y_{\backslash k} R_k'X_{k} dA_k \right).
\end{align}
The differential $dA_k$ appearing in \eqref{eqn:dR} can be easily computed starting from the definition of the definition of $A_k$. Specifically, we have:
\begin{align}
    dA_k &= d \left(X_{\backslash k}'X_{\backslash k} + N_T\Lambda\Lambda\right)^{-1} \nonumber \\
    &= -N_T A_k d\left(\Lambda\Lambda\right) A_k,\label{eqn:dA2}
\end{align}
where the following identity is used:
\begin{align} \label{eqn:lemma1}
d\left(W^{-1}\right) = -W^{-1} \left(dW\right) W^{-1} \,,
\end{align}
for any invertible square matrix $W \in \mathbb{R}^{m \times m}$.
By substituting \eqref{eqn:dA2} into \eqref{eqn:dLk}, we obtain:
\begin{align} \label{eqn_dLk3}
dL_k &= - \frac{N_T}{N_V} \mathrm{tr}\left[X_{\backslash k}'Y_{\backslash k} R_k'X_{k} A_k d\left(\Lambda\Lambda\right) A_k \right] \nonumber \\
&= - \frac{N_T}{N_V} \mathrm{tr}\left[A_k X_{\backslash k}'Y_{\backslash k} R_k'X_{k} A_k d\left(\Lambda\Lambda\right)\right] \nonumber \\
&= - \frac{N_T}{N_V} \mathrm{tr}\left[\hat \bw^{(\backslash k)} R_k'X_{k} A_k d\left(\Lambda\Lambda\right) \right] \,.
\end{align}
Finally, from the product rule  and by defining $B_k=A_k' X_k' R_k \left(\hat \bw^{(\backslash k)}\right)'$, \eqref{eqn_dLk3} can be written as:
\begin{align}
dL_k &= - \frac{N_T}{N_V} \mathrm{tr}\Big[\hat \bw^{(\backslash k)} R_k'X_{k} A_k \left(d\Lambda\right) \Lambda\nonumber\\&\qquad + \hat \bw^{(\backslash k)} R_k'X_{k} A_k \Lambda d\Lambda \Big] \nonumber \\
&= - \frac{N_T}{N_V} \mathrm{tr}\left[B_k' \left(d\Lambda\right) \Lambda + B_k' \Lambda d\Lambda \right] \nonumber \\
&= - \frac{N_T}{N_V} \mathrm{tr}\left[\Lambda B_k' \left(d\Lambda\right)  + B_k' \Lambda d\Lambda \right] \nonumber \\
&= - \frac{N_T}{N_V} \mathrm{tr}\left[\left(\Lambda B_k' + B_k' \Lambda \right) d\Lambda \right] \label{eqn:dLk_final} \,.
\end{align}
Using the following property: for any matrix $U \in \mathbb{R}^{m \times n}$ and  $V \in \mathbb{R}^{m \times p}$:
\begin{align}
&\mathrm{tr}\left( U'V \right) = \mathrm{vec}\left( U \right)' \mathrm{vec}\left( V \right) \,, \label{eqn:lemma2a}
\end{align}
\eqref{eqn:dLk_final} becomes:
\begin{align}
dL_k &= - \frac{N_T}{N_V} \mathrm{vec}\left(\Lambda B_k + B_k \Lambda \right)' d\mathrm{vec}\Lambda. \label{eq:diff_eval_criterion_final_Lambda}
\end{align}
By combining \eqref{eqn:diff_eval_criterion} and \eqref{eq:diff_eval_criterion_final_Lambda}, the differential $dE$ can be  written as:
\begin{align} \label{eqn:dE_secondproof}
    dE = - \frac{N_T}{KN_V}  \sum_{k=1}^K  \mathrm{vec}\left(\Lambda B_k + B_k \Lambda \right)' d\mathrm{vec}\Lambda.
\end{align}
Thus, from \eqref{eqn:dE_secondproof} and \eqref{eqn:first_identification}, we observe:
\begin{align}
\nabla_{\Lambda} E &= - \frac{N_T}{K N_V} \sum_{k=1}^K \mathrm{vec}\left(\Lambda B_k + B_k \Lambda \right),
\end{align}
which  proves \eqref{eqn:gradient_eval_criterion_Lambda}.
In order to prove \eqref{eqn:gradient_eval_criterion_lambda}, we rewrite  \eqref{eqn:dLk_final} as:
\begin{align} \label{eqn:dL_klambda}
dL_k &= - \frac{N_T}{N_V} \mathrm{diag}\left(\Lambda B_k + B_k \Lambda \right)' d\lambda,
\end{align}
where we used the following properties: 
\begin{subequations} \label{eqn:app:lemma}
\begin{align}
&\mathrm{tr}\left( W \mathrm{Diag}\left(v \right) \right) = \mathrm{diag}\left(W \right)' v \,, \label{eqn:lemma2b}\\
&d\mathrm{Diag}(v) = \mathrm{Diag}(dv) \,,\label{eqn:lemma2c}
\end{align}
\end{subequations}
for any square matrix   $W \in \mathbb{R}^{m \times m}$ and vector $v \in \mathbb{R}^{m}$.
By combining \eqref{eqn:diff_eval_criterion} and \eqref{eqn:dL_klambda}, the differential $dE$ can be  written as:
\begin{align} \label{eqn:dE_thirdproof}
    dE = - \frac{N_T}{KN_V}  \sum_{k=1}^K  \mathrm{diag}\left(\Lambda B_k + B_k \Lambda \right)' d\lambda.
\end{align}
Thus, from \eqref{eqn:dE_thirdproof} and \eqref{eqn:first_identification_lambda}, we observe:
\begin{align}
\nabla_{\lambda} E &= - \frac{N_T}{K N_V} \sum_{k=1}^K \mathrm{diag}\left(\Lambda B_k + B_k \Lambda \right),
\end{align}
which proves \eqref{eqn:gradient_eval_criterion_lambda}.
\section{Derivation of \eqref{eq:grad_reg_criterion_final_Lambda} and \eqref{eq:grad_reg_criterion_final_lambda}}
\label{sec:appendix_grad} \label{app-der2}
First, let us rewrite the regularization term $\Q$ in \eqref{eqn:Q} as:
\begin{align} \label{eqn:Q_app}
\Q =  \sum_{k=1}^K \Q_k,
\end{align}
with:
\begin{align} \label{eqn:regularization_in_validation_term}
\Q_k & = \frac{1}{2} \mu \left\|  \Lambda \hat \bw^{(\backslash k)} \right\|_F^2 = \frac{1}{2} \mu \, \mathrm{tr} \left(\left(\hat \bw^{(\backslash k)}\right)' \Lambda \Lambda \hat \bw^{(\backslash k)} \right) \nonumber \\
&= \frac{1}{2} \mu \, \mathrm{tr} \left(D_k' D_k \right),
\end{align}
where $D_k = \Lambda \hat \bw^{(\backslash k)}$ and the dependence of $\bw^{(\backslash k)}$ on $\Lambda$ is omitted to ease reading. 
Differentiating both sides of \eqref{eqn:regularization_in_validation_term} and using the trace and differential operator properties already adopted in Appendix \ref{app-der1} we obtain:
\begin{align} \label{eqn:regularization_in_validation_term1}
dQ_k &= \frac{1}{2} \mu \, d\left[ \mathrm{tr} \left(D_k' D_k \right) \right]  = \frac{1}{2} \mu \, \mathrm{tr} \left[ d \left(D_k' D_k \right) \right] \nonumber \\
&= \frac{1}{2} \mu \, \mathrm{tr} \left[ \left(dD_k'\right)D_k + D_k'dD_k \right] \nonumber \\
&= \frac{1}{2} \mu \, \mathrm{tr} \left[ \left(dD_k\right)'D_k + D_k'dD_k \right] \nonumber \\
& = \frac{1}{2} \mu \, \mathrm{tr} \left( D_k'dD_k + D_k'dD_k \right) \nonumber \\
&= \mu \, \mathrm{tr} \left( D_k'dD_k \right).
\end{align}
From the definition of $D_k$, \eqref{eqn:regularization_in_validation_term1} can be rewritten as:
\begin{align} \label{eqn:dQk_2}
dQ_k & = \mu \, \mathrm{tr} \left[ D_k'd\left(\Lambda \hat \bw^{(\backslash k)}  \right) \right] \nonumber \\
&= \mu \, \mathrm{tr} \left[ D_k'\left(d\Lambda \right) \hat \bw^{(\backslash k)}  + D_k' \Lambda d\hat \bw^{(\backslash k)}  \right] \nonumber \\
&= \mu \left[ \mathrm{tr}\left(\hat \bw^{(\backslash k)}  D_k' d\Lambda \right) +  \mathrm{tr}\left(D_k' \Lambda d\hat \bw^{(\backslash k)}  \right)\right].
\end{align}
Let us now focus on the second term in the right side of \eqref{eqn:dQk_2}, which is rewritten by using the same properties adopted in Appendix \ref{app-der1}, as: 
\begin{align}
 &\mathrm{tr}\left(D_k' \Lambda d\hat \bw^{(\backslash k)} \right) = \nonumber\\
&= \mathrm{tr} \left[D_k' \Lambda d\left(\left(X_{\backslash k}'X_{\backslash k} + N_T\Lambda\Lambda\right)^{-1}X_{\backslash k}'Y_{\backslash k} \right) \right] \nonumber\\
&= \mathrm{tr} \left[ D_k' \Lambda \left(dA_k\right)X_{\backslash k}'Y_{\backslash k} \right] \nonumber\\
&= \mathrm{tr} \left(X_{\backslash k}'Y_{\backslash k} D_k' \Lambda dA_k \right) \nonumber\\
&= -N_T \, \mathrm{tr} \left[X_{\backslash k}'Y_{\backslash k} D_k' \Lambda A_k d\left(\Lambda\Lambda\right)A_k  \right] \nonumber\\
&= -N_T \, \mathrm{tr} \left[\hat \bw^{(\backslash k)} D_k' \Lambda A_k d\left(\Lambda\Lambda\right)  \right] \nonumber\\
&= -N_T \, \mathrm{tr} \left[\hat \bw^{(\backslash k)} D_k' \Lambda A_k \left(d\Lambda\right)\Lambda +  \hat \bw^{(\backslash k)} D_k' \Lambda A_k \Lambda d\Lambda \right]. \nonumber\\
\end{align}
By defining  $G_k=A_k'\Lambda'D_k \left(\hat \bw^{(\backslash k)}\right)'$, the above equation can be written in the compact form:
\begin{align} \label{eqn:regularization_in_validation_term2}
\mathrm{tr}\left(D_k' \Lambda d\hat \bw^{(\backslash k)} \right) &= -N_T \, \mathrm{tr} \left[G_k' \left(d\Lambda\right)\Lambda + G_k' \Lambda d\Lambda \right] \nonumber\\
&= -N_T \, \mathrm{tr} \left[\left(\Lambda G_k' + G_k' \Lambda \right) d\Lambda \right]. \nonumber\\
\end{align}
Then, by substituting \eqref{eqn:regularization_in_validation_term2} into \eqref{eqn:dQk_2}, we obtain: 
\begin{align} \label{eqn:regularization_in_validation_term3}
dQ_k &= \mu \left[ \mathrm{tr}\left(\hat \bw^{(\backslash k)} D_k' d\Lambda \right) -N_T \, \mathrm{tr} \left(\left(\Lambda G_k' + G_k' \Lambda \right) d\Lambda \right) \right] \nonumber\\
&= \mu \left[ \mathrm{tr}\left(\left(\hat \bw^{(\backslash k)} D_k' -N_T \left(\Lambda G_k' + G_k' \Lambda \right)\right) d\Lambda \right) \right]. 
\end{align}
Thanks to \eqref{eqn:lemma2a}, \eqref{eqn:regularization_in_validation_term3} can be written as:
\begin{align} \label{eqn:regularization_in_validation_term4}
d\Q_k &= \mu \, \mathrm{vec}\left( D_k \hat \bw^{(\backslash k)} - N_T \left(\Lambda G_k + G_k \Lambda \right)\right)' d \mathrm{vec} \Lambda. \end{align}
Based on the same considerations discussed in Appendix \ref{app-der1}, from \eqref{eqn:regularization_in_validation_term4}, and the definition of $\Q$ in \eqref{eqn:Q}, we obtain:
\begin{align}
\nabla_{\Lambda}\Q &= \mu \sum_{k=1}^K \mathrm{vec}\left( D_k \left(\hat \bw^{(\backslash k)}\right)' - N_T \left(\Lambda G_k + G_k \Lambda \right)\right),
\end{align}
thus proving \eqref{eq:grad_reg_criterion_final_Lambda}.
Using the matrix properties in \eqref{eqn:app:lemma}, \eqref{eqn:regularization_in_validation_term3} can be also written as:
\begin{align}
d\Q_k &= \mu \, \mathrm{diag} \left( D \left(\hat \bw^{(\backslash k)}\right)' - N_T \left(\Lambda G_k + G_k \Lambda \right) \right)' d\lambda.
\end{align}
Therefore:
\begin{align}
\nabla_{\lambda}\Q &= \mu \sum_{k=1}^K \mathrm{diag} \left( D_k \left(\hat \bw^{(\backslash k)}\right)' - N_T \left(\Lambda G_k + G_k \Lambda \right) \right),
\end{align}
thus proving \eqref{eq:grad_reg_criterion_final_lambda}.
\end{document}